\documentclass[pmlr,twoside,11pt]{article}
\usepackage{jmlr2e}

\usepackage{xargs}                      
\usepackage{xcolor}  
\usepackage[colorinlistoftodos,prependcaption,textsize=small]{todonotes}
\newcommandx{\unsure}[2][1=]{\todo[linecolor=red,backgroundcolor=red!25,bordercolor=red,#1]{#2}}
\usepackage{graphicx}
\usepackage{hyperref} 
\usepackage{url}
\usepackage{amssymb}
\usepackage{bm}
\usepackage{algorithm}
\usepackage{algpseudocode}
\usepackage{caption}
\usepackage{subcaption}
\usepackage{wrapfig}
\usepackage{multirow}
\usepackage{amsmath}
\usepackage{marginnote}
\usepackage{amsmath}
\usepackage{natbib} 

\jmlrheading{x}{2018}{x}{5/15}{x}{Natasha Jaques, Jennifer McCleary, Jesse Engel, David Ha, Fred Bertsch, Douglas Eck, Rosalind Picard}

\ShortHeadings{Improving a deep generative sketching model with facial feedback}{Jaques, McCleary, Engel, Ha, Bertsch, Eck, and Picard}
\firstpageno{1}

\begin{document}

\title{Learning via social awareness: Improving a deep generative sketching model with facial feedback}

\author{\name Natasha Jaques$^{12}$ \email jaquesn@mit.edu \\
		\name Jennifer McCleary$^1$ \email mccleary@mit.edu \\
        \name Jesse Engel$^2$ \email jesseengel@google.com \\
        \name David Ha$^2$ \email hadavid@google.com \\
        \name Fred Bertsch$^2$ \email fredbertsch@google.com \\
        \name Douglas Eck$^2$ \email deck@google.com \\
        \name Rosalind Picard$^1$ \email picard@mit.edu \\\\
       \addr $^1$Media Lab, Massachusetts Institute of Technology, Cambridge, MA 02139, USA \\
       \addr $^2$Google Brain, Mountain View, CA 94043, USA}

\editor{} 

\maketitle

\begin{abstract}
A known deficit of modern machine learning (ML) and deep learning (DL) methodology is that models must be carefully fine-tuned in order to solve a particular task. Most algorithms cannot generalize well to even highly similar tasks, let alone exhibit signs of artificial general intelligence (AGI). To address this problem, researchers have explored developing loss functions that act as intrinsic motivators that could drive an ML or DL agent to learn across a number of domains. This paper argues that an important and useful intrinsic motivator is that of social interaction. We posit that making an AI agent aware of implicit social feedback from humans can allow for faster learning of more generalizable and useful representations, and could potentially impact AI safety. 
We collect social feedback in the form of facial expression reactions to samples from Sketch RNN, an LSTM-based variational autoencoder (VAE) designed to produce sketch drawings. We use a Latent Constraints GAN (LC-GAN) to learn from the facial feedback of a small group of viewers, by optimizing the model to produce sketches that it predicts will lead to more positive facial expressions. We show in multiple independent evaluations that the model trained with facial feedback produced sketches that are more highly rated, and induce significantly more positive facial expressions. Thus, we establish that implicit social feedback can improve the output of a deep learning model.
\end{abstract}


\section{Introduction}
Despite the recent rapid and compelling progress in ML and DL, modern AI is still remarkably far from approximating the intelligence of even simple animals.  A notable deficit is the degree of explicit supervision required in order to learn, either through labeled samples or well-defined external rewards such as points in a game. The limited scope of such supervision will not enable the development of a generally intelligent AI.

For this reason, some researchers have focused on intrinsic motivators, inherent drives that cause the agent to learn representations that are useful across a variety of tasks and environments.  Examples include curiosity (a drive for novelty) \citep{pathak2017curiosity}, and empowerment (a drive for the ability to manipulate the environment) \citep{capdepuy2007maximization}. However, so far this research has overlooked an important intrinsic motivator for humans: the drive for positive social interactions. 

We argue that making an AI agent intrinsically motivated to obtain a positive social reaction from humans in its environment is an important new research direction. Specifically, the agent should be able to recognize implicit feedback from humans in the form of facial expressions, body language, or tone in voice and text, and optimize for actions that appear to please humans as measured through these signals. 

The representations learned by such an agent are more likely to capture dimensions of the task that are relevant to human satisfaction. This has meaningful implications for questions of AI safety; an AI agent motivated by satisfaction expressed by humans will be less likely to take actions against human interest. Such an agent will also be better suited to perform tasks which already involve AI. Imagine if a home assistant could sense when a user responds with an angry or frustrated tone and this acted as a negative incentive, training the algorithm not to repeat the action that led to the user's frustration? Rather than requiring the user to manually train the device, it could learn quickly through passive sensing of the user's emotional state, leading to a more immediately satisfying experience for the user. Finally, some machine learning problems --- including the one under investigation in this paper --- cannot be solved without human feedback; when the objective function is human aesthetic preference, it cannot be approximated without human input.

Social awareness may be a key component of AGI. There is substantial evidence that emotion recognition, which is critical for empathy and successful social interaction, plays an influential role in cognitive development in humans \citep{kujawa2014emotion}. According to Social Learning Theory \citep{bandura1977social}, observing the attitudes and behaviors of others is a central component of how humans learn both intelligent behavior and how to adapt to new situations. It has been argued that social learning is responsible for the rapid cultural evolution of the human species \citep{van2011social}. Given the importance of cultural evolution to humans' technological success, endowing a deep learning agent with the ability to perceive and benefit from this socially exchanged cultural knowledge could allow it to rapidly develop more generalizable knowledge representations.


In this work we demonstrate the utility of learning through implicit social feedback via an experiment in which samples generated by a deep learning model are presented to people, and their facial expression response is detected. The model is Sketch RNN \citep{ha2017neural}, an LSTM-based VAE with a Mixture Density Network
output, designed to produce sketch drawings. Using a newly developed technique known as Latent Constraints \citep{engel2017latent}, we train a Generative Adversarial Network (GAN) to produce VAE embedding vectors that, when decoded by Sketch RNN, are more likely to produce drawings that lead to positive facial expressions such as smiling. In a rigorous, double-blind evaluation, we show that samples from the social feedback model generate statistically significantly better affective responses than the prior, and are consistently rated as more preferred by human judges. Thus, this experiment is a first step in demonstrating that deep learning models are able to improve in quality as a result of learning from implicit social feedback. 

\section{Related work}
Many affective computing papers have addressed how to automatically detect facial expressions (e.g. \cite{senechal2015facial}). A comprehensive review of this work is out of scope for this paper. We instead build on this work by assuming that an accurate facial expression detector is already available, and asking what can be learned using this facial feedback.

Previous work has attempted to train ML and DL models to approximate human preferences. For example, \citet{knox2009interactively} ask users to press a button to teach a reinforcement learning (RL) model to play \textit{Mountain Car}, and model human reaction latencies as a Gamma distribution in order to distribute the reward appropriately over past time steps. A more recent work attempts to train a deep learning model from human preferences, by first training an approximator of human button presses using supervised learning, and then using this to train an RL model \citep{christiano2017deep}. However, both of these approaches  require the human to provide explicit supervision by manually entering feedback. In contrast, our approach enables learning from implicit social cues that can be obtained ubiquitously, through awareness of the non-verbal reactions people naturally provide. Essentially, we obtain human-in-the-loop training without additional human effort.

The closest work to our own of which we are aware is an approach that used valence and engagement, detected via facial expressions, as a reward function in a Q-learning framework \citep{gordon2016affective}. The goal of the project was to allow an intelligent tutoring system to adapt its behavior so that children would remain engaged while using the system. While this is an excellent example of learning from implicit social feedback, the goals of this paper are quite distinct from our own. We believe we are the first authors to use implicit social feedback to improve a generative deep learning model. Our model attempts to learn to improve its ability to produce creative content by observing the implicit responses it receives from human judges. This process could be considered analogous to a human artist fine-tuning her work after she observes critics' reactions. 

\section{Methods}
\subsection{Study design}
To gather social feedback, we focused on facial expression recognition, since this is currently one of the most reliable and accurate ways to detect social signals \citep{senechal2015facial}. The facial expression detector employed for this project is a pre-trained convolutional network trained to detect common facial expressions.

To obtain facial feedback at scale, we built a web app that serves samples from a deep learning model while recording the user's facial expressions with a webcam. The webcam images were fed into the facial-expression detection network and used to compute the intensities of common expressions, including amusement, contentment, surprise, sadness, and concentration. Due to the degree of inter-individual variation in users' resting facial expression, these intensities were normalized against each user's average expression to produce value vectors $v$. The app is also capable of collecting Likert-scale ratings of sketch quality and asking users to choose which of two sketches they prefer. These mechanisms were used to collect evaluation data. The app can be viewed at \url{https://facial-feedback-for-ai.appspot.com/}.

To test the hypothesis that facial feedback can improve the outputs of a deep learning model, we sought a model for which the outputs were likely to generate a natural facial expression response. We chose Sketch RNN \citep{ha2017neural}, a model which generates sequences of strokes that form a sketched image of a common object, vehicle, or animal (see Figure \ref{fig:sketches}). Such sketches were determined to elicit facial responses in initial tests. 

\subsection{Machine learning techniques}
Sketch RNN is a VAE that was trained in an unsupervised manner on a large corpus on human sketch data collected via \emph{Quick, Draw!}\footnote{\url{https://quickdraw.withgoogle.com/}}. The sketches are represented as sequences of coordinates that represent the points where the pen is placed during sketching. The architecture of Sketch RNN comprises: a) a bidirectional LSTM encoder that projects each input sketch into a latent embedding vector $z$, b) an LSTM decoder which takes $z$ as input and generates a sequence of parameters for c) a Gaussian Mixture Model that generates the $(x,y)$ coordinates of the tip of the pen during each stroke. This Mixture Density Network (MDN) approach is similar to prior work on handwriting generation \citep{graves2013generating}. 

The design of Sketch RNN provides important benefits that facilitate optimizing the model with facial feedback. First, due to the variational constraint, it is straightforward to sample a latent vector $z \sim \mathcal{N}(0,I)$ and feed this into the Sketch RNN decoder to produce a recognizable sketch. Second, the latent embeddings $z$ learned by Sketch RNN provide a clean, compressed representation of sketch drawings. These features allowed us to apply a newly developed technique known as Latent Constraints \citep{engel2017latent} in order to learn to produce sketches likely to lead to positive facial expressions.

\begin{figure}[h]
\begin{center}
\includegraphics[width=.55\linewidth]{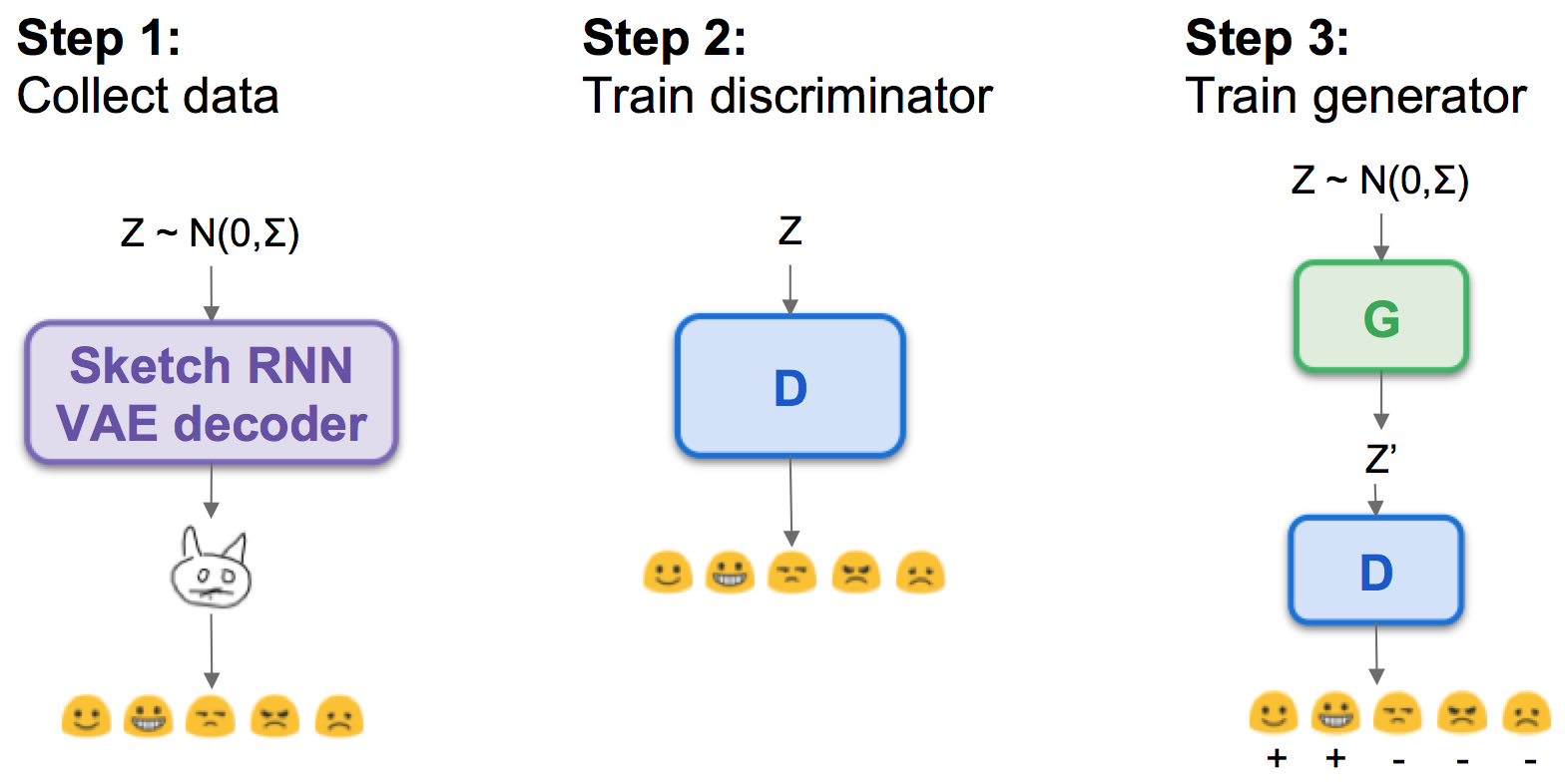}
\end{center}
\caption{Steps involved in training the LC-GAN facial feedback model.}
\label{fig:model_diagram}
\vspace{-0.2cm}
\end{figure}

The latent constraints GAN (LC-GAN) is a GAN applied to the latent embedding space of a VAE. The steps of training this model to use facial feedback are shown in Figure \ref{fig:model_diagram}. We first sample a number of $z$ vectors from the VAE prior ($\mathcal{N}(0,I)$), and feed these into the Sketch RNN decoder to obtain sketches. These sketches are shown to users, and the intensity of their facial expression responses is recorded; we refer to these intensities as the \emph{value} of a sketch, $v$. The $z$ vectors are then used as input to a discriminator $D(z) \rightarrow v$, which is trained to estimate the value $v$ of different regions of the latent space; for example, which regions decode to sketches that produced the highest intensity of smiles. A generator $G(z) \rightarrow z'$ is then trained to convert a randomly sampled $z$ into a modified $z'$ that produces a higher $v$. In fact, the generator uses a gating mechanism to control how heavily the original $z$ is modified. The generator loss is $\mathcal{L}_G=-\log D(z')$. Because the Sketch RNN latent space is a compressed, 128-dimensional, robust representation, the discriminator is able to learn a value function on $z$ even with relatively small sample sizes. 

\section{Experiments}
Data collection for the experiments was conducted in four phases. In the initial phase, we ran a pilot study on $7$ users who viewed a total of $30$ sketches, in which we collected both facial expressions and Likert-scale ratings of sketch quality. Then, we used the webapp to collect facial reactions from 28 users to a total of 334 sketches, recording the embedding vector $z$ for each sketch. These $(z,v)$ pairs were used to train the LC-GAN. Finally, two phases of data collection were used to evaluate the model. Both involved rigorous, double-blind experiments in which we randomly generated hundreds of samples from both the facial feedback and baseline models, and displayed them in random order to users ``in the wild'', using their personal webcams, without experimenter supervision. The first evaluation sought to establish that sketches from the LC-GAN are able to elicit more positive facial expressions than the original Sketch RNN. For this test, we obtained evaluation data from $76$ users, spanning 536 sketches. The second evaluation asked users to rate which of two sketches they preferred; we collected 4,692 ratings from 79 users. 

\begin{figure}
\centering
\begin{subfigure}{.5\textwidth}
  \centering
  \includegraphics[width=.65\linewidth]{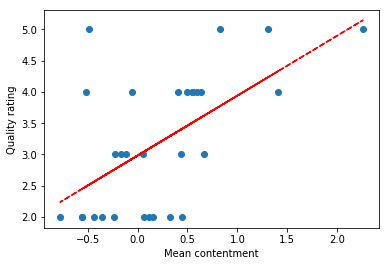}
  \caption{Contentment, $r$=-.58, $p<$.001}
  \label{fig:sub1}
\end{subfigure}%
\begin{subfigure}{.5\textwidth}
  \centering
  \includegraphics[width=.65\linewidth]{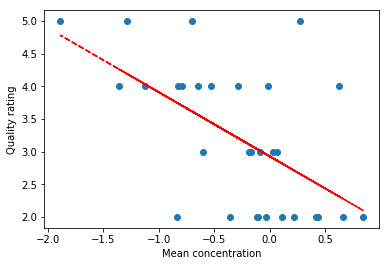}
  \caption{Concentration, $r$=.58, $p<$.001}
  \label{fig:sub2}
\end{subfigure}
\caption{Correlation between quality ratings and expressions.}
\label{fig:ux_corr}
\vspace{-0.2cm}
\end{figure}

\section{Results}
\subsection{Facial expression analysis}
Optimizing for user preference requires knowing which facial expressions indicate that the user likes a sketch. Therefore, we used the pilot study data to assess how users' ratings of sketch quality related to their facial expressions. We found significant positive correlations with contentment and amusement (smiling), and significantly negative correlations with sadness and concentration (frowning), as shown in Table \ref{tab:corrs}; examples are shown in Figure \ref{fig:ux_corr}. Notably, these results indicate that implicit facial feedback carries an informative signal about the user's preferences. 

\begin{wraptable}{l}{55mm}
  \caption{Correlations between expressions and quality ratings.}
  \label{tab:corrs}
    \begin{tabular}{lll}
  \textbf{Emotion metric} & \textbf{$r$}     & \textbf{$p$}    \\ \hline
  Contentment        & .582           & .001          \\
  Amusement          & .546           & .002  \\
  Concentration      & -.576          & .001          \\
  Sadness            & -.405		  & .026 \\
  \end{tabular}
\end{wraptable}

However, learning from facial feedback still represents a challenging problem; there is a high degree of inter-individual variability, the meaning of facial expressions may be extremely context dependent, and the data can be remarkably noisy. In addition to noise introduced through inaccuracies in the detector, there are many confounding reasons that may cause a person to make a given facial expression. 
\begin{wrapfigure}{R}{.45\textwidth}
\centering
\vspace{-0.3cm}
\includegraphics[width=.45\textwidth]{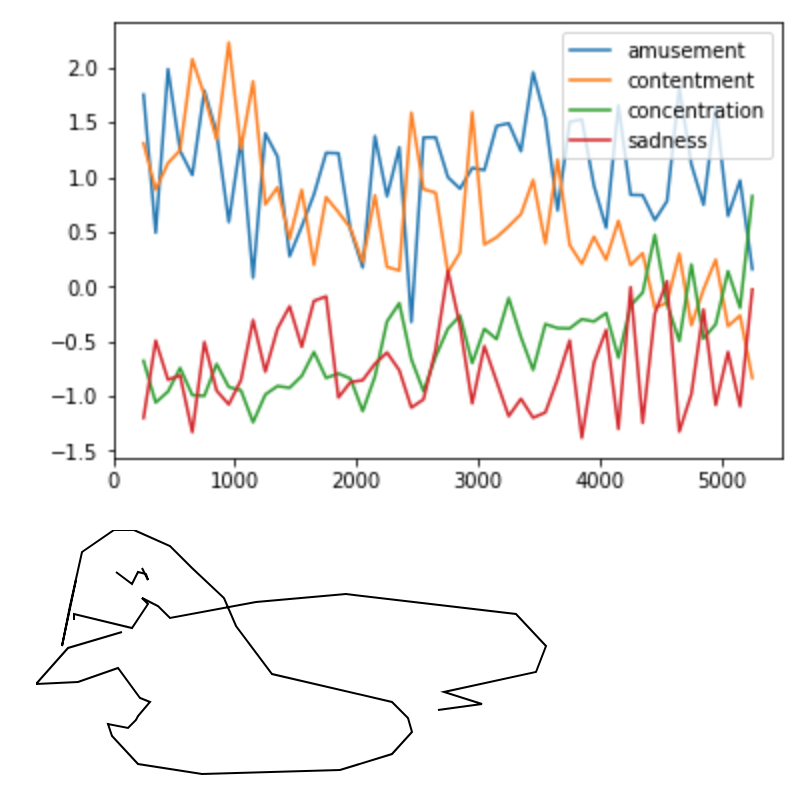}
\caption{\label{fig:scribble}The user expresses contentment and amusement at a scribbled sketch.}
\vspace{-0.3cm}
\end{wrapfigure}
For example, \cite{hoque2011acted} found that users tend to smile when they are frustrated. In our case, we noticed that users tend to smile simply at the concept of an AI making drawings, or even at their own face as shown to them via the webcam feed. This can lead to highly misleading interactions; Figure \ref{fig:scribble} shows an example in which the user smiles profusely at a drawing that is no better than a scribble. 

Finally, the difficulty of modeling user preference through facial expressions is enhanced by the non-stationarity of the data. We found that users' facial expressions tended to change over repeated interactions with the system. There were significant relationships between the number of previous sketches viewed by the user and the user's average sadness ($r(751) = 0.248, p<.001$) and concentration $r(751) = -0.158, p<.001$). Thus, the meaning of an intense expression of concentration may change depending on when it occurs in the user's interaction with the app.

\subsection{Machine learning results}

\begin{wrapfigure}{L}{.5\textwidth}
\centering
\vspace{-0.3cm}
\includegraphics[width=.5\textwidth]{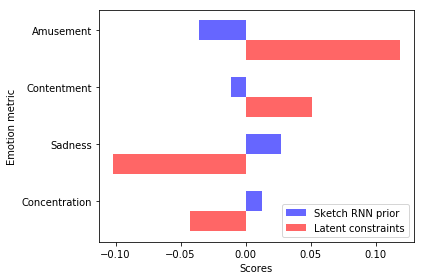}
\caption{\label{fig:bar}Sketches from the LC-GAN generate more positive expressions.}
\vspace{-0.3cm}
\end{wrapfigure}

\begin{figure}[H]
\begin{center}
\includegraphics[width=\linewidth]{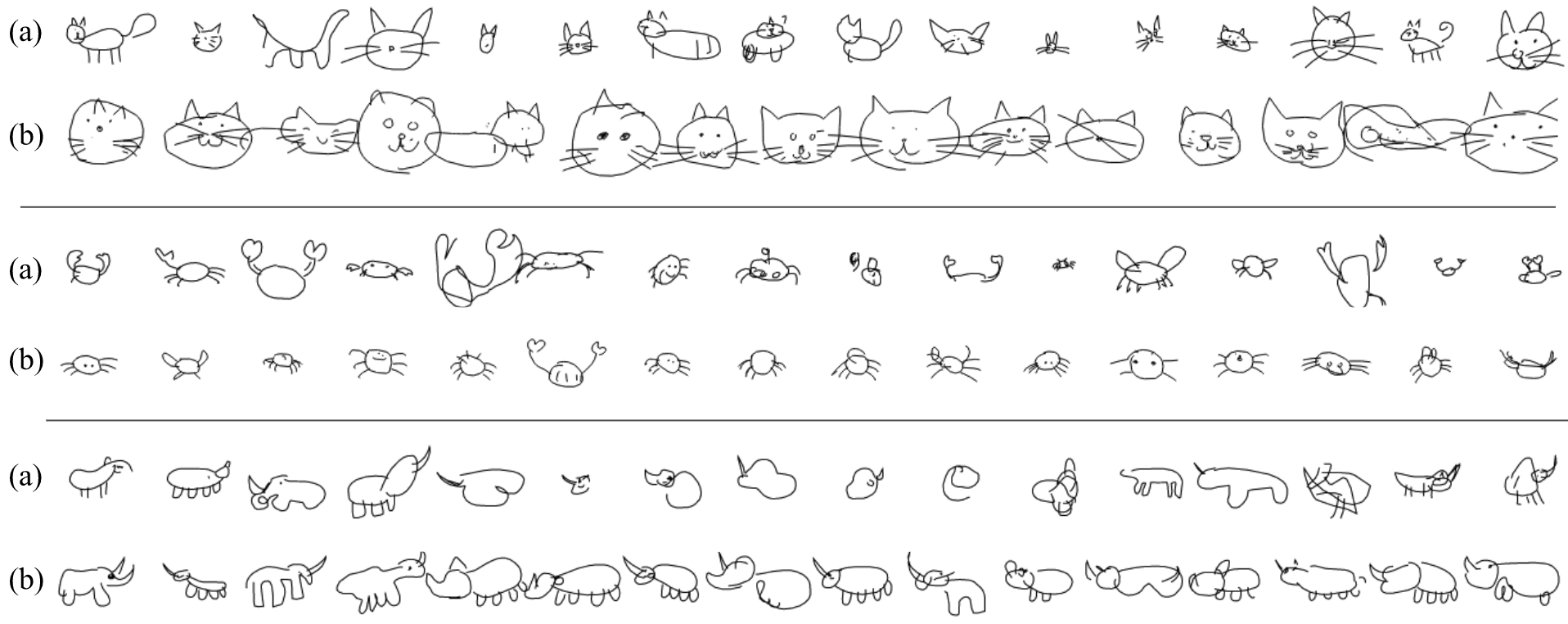}
\end{center}
\caption{Samples drawn randomly from the cat, crab, and rhinoceros sketch classes, produced by (a) the original Sketch RNN, and (b) the LC-GAN trained on a small amount of social feedback.}
\label{fig:sketches}
\vspace{-0.2cm}
\end{figure}

In spite of the noise and non-stationarity inherent in the data, we found that the LC-GAN was able to use the facial feedback to learn produce significantly higher quality sketches. Given the direction of the relationships discovered between facial expressions and quality, we trained the LC-GAN to maximize amusement and contentment, and minimize concentration and sadness. Although relatively little data was collected (63-69 samples per sketch class), the LC-GAN was able to effectively optimize for more pleasing sketches. Figure \ref{fig:sketches} shows the difference between samples produced with the Sketch RNN prior and the LC-GAN. The LC-GAN appears to have learned that people smile more and frown less for cats with larger, smiling faces with whiskers. Similarly, the quality of crab and rhinoceros sketches generated by the LC-GAN appears to be consistently higher. For example, the original rhinoceros model often produced sketches that did not resemble a rhinoceros, or were no better than scribbles. After training with a small amount of facial feedback, the LC-GAN model consistently produces more realistic drawings.

The evaluation data revealed that humans found sketches from the LC-GAN model to be significantly better. In the first experiment, we were able to support the hypothesis that sketches from the LC-GAN model generate significantly more positive facial expressions than the original Sketch RNN. Figure \ref{fig:bar} shows the results of this evaluation, indicating that all of the facial expression metrics improved in the expected direction under the LC-GAN. Two of the metrics reached statistical significance: mean amusement, $t(535)=2.31, p<.05$, and mean sadness, $t(535)=-2.01, p<.05$.
In the second part of the evaluation, we tested the hypothesis that humans actually rate the quality of the LC-GAN sketches as higher. 
Users reported preferred the LC-GAN 2843 times, as opposed to 1770 for the original Sketch RNN, a significant improvement as shown in a Binomial test, $p < .0001$. 


\section{Conclusions and Future Work}
We have demonstrated that implicit social feedback in the form of facial expressions not only can reflect user preference, but also can significantly improve the performance of a deep learning model.

There are many ways to enhance and extend this work. For example, we could use an RL framework to improve the model's ability to draw based on facial feedback. Further, our current model makes no use of the evolving temporal dynamics of the collected facial expressions, instead relying on an average intensity over viewing the sketch. A more sophisticated system could use the alignment between the process of drawing the sketch and the user's expressions to gain better temporal credit assignment in an RL framework.

{\footnotesize
\acks{We would like to thank James Tolentino, Ira Blossom, Adrien Baranes, Katherine Lee, Chris Han, Curtis Hawthorne, Rebecca Salois, Josh Lovejoy, Sherol Chen, and Mike Dory for their contributions to this project.}

\bibliography{facial_feedback}}

\begin{thebibliography}{13}
\providecommand{\natexlab}[1]{#1}
\providecommand{\url}[1]{\texttt{#1}}
\expandafter\ifx\csname urlstyle\endcsname\relax
  \providecommand{\doi}[1]{doi: #1}\else
  \providecommand{\doi}{doi: \begingroup \urlstyle{rm}\Url}\fi

\bibitem[Bandura and Walters(1977)]{bandura1977social}
Albert Bandura and Richard~H Walters.
\newblock Social learning theory.
\newblock 1977.

\bibitem[Capdepuy et~al.(2007)Capdepuy, Polani, and
  Nehaniv]{capdepuy2007maximization}
Philippe Capdepuy, Daniel Polani, and Chrystopher~L Nehaniv.
\newblock Maximization of potential information flow as a universal utility for
  collective behaviour.
\newblock In \emph{Artificial Life, 2007. ALIFE'07. IEEE Symposium on}, pages
  207--213. Ieee, 2007.

\bibitem[Christiano et~al.(2017)Christiano, Leike, Brown, Martic, Legg, and
  Amodei]{christiano2017deep}
Paul~F Christiano, Jan Leike, Tom Brown, Miljan Martic, Shane Legg, and Dario
  Amodei.
\newblock Deep reinforcement learning from human preferences.
\newblock In \emph{Advances in Neural Information Processing Systems}, pages
  4302--4310, 2017.

\bibitem[Engel et~al.(2017)Engel, Hoffman, and Roberts]{engel2017latent}
Jesse Engel, Matthew Hoffman, and Adam Roberts.
\newblock Latent constraints: Learning to generate conditionally from
  unconditional generative models.
\newblock \emph{arXiv preprint arXiv:1711.05772}, 2017.

\bibitem[Gordon et~al.(2016)Gordon, Spaulding, Westlund, Lee, Plummer,
  Martinez, Das, and Breazeal]{gordon2016affective}
Goren Gordon, Samuel Spaulding, Jacqueline~Kory Westlund, Jin~Joo Lee, Luke
  Plummer, Marayna Martinez, Madhurima Das, and Cynthia Breazeal.
\newblock Affective personalization of a social robot tutor for children's
  second language skills.
\newblock In \emph{AAAI}, pages 3951--3957, 2016.

\bibitem[Graves(2013)]{graves2013generating}
Alex Graves.
\newblock Generating sequences with recurrent neural networks.
\newblock \emph{arXiv preprint arXiv:1308.0850}, 2013.

\bibitem[Ha and Eck(2017)]{ha2017neural}
David Ha and Douglas Eck.
\newblock A neural representation of sketch drawings.
\newblock \emph{arXiv preprint arXiv:1704.03477}, 2017.

\bibitem[Hoque and Picard(2011)]{hoque2011acted}
Mohammed Hoque and Rosalind~W Picard.
\newblock Acted vs. natural frustration and delight: Many people smile in
  natural frustration.
\newblock In \emph{Automatic Face \& Gesture Recognition and Workshops (FG
  2011), 2011 IEEE International Conference on}, pages 354--359. IEEE, 2011.

\bibitem[Knox and Stone(2009)]{knox2009interactively}
W~Bradley Knox and Peter Stone.
\newblock Interactively shaping agents via human reinforcement: The tamer
  framework.
\newblock In \emph{Proceedings of the fifth international conference on
  Knowledge capture}, pages 9--16. ACM, 2009.

\bibitem[Kujawa et~al.(2014)Kujawa, Dougherty, Durbin, Laptook, Torpey, and
  Klein]{kujawa2014emotion}
Autumn Kujawa, LEA Dougherty, C~Emily Durbin, Rebecca Laptook, Dana Torpey, and
  Daniel~N Klein.
\newblock Emotion recognition in preschool children: Associations with maternal
  depression and early parenting.
\newblock \emph{Development and psychopathology}, 26\penalty0 (1):\penalty0
  159--170, 2014.

\bibitem[Pathak et~al.(2017)Pathak, Agrawal, Efros, and
  Darrell]{pathak2017curiosity}
Deepak Pathak, Pulkit Agrawal, Alexei~A Efros, and Trevor Darrell.
\newblock Curiosity-driven exploration by self-supervised prediction.
\newblock \emph{arXiv preprint arXiv:1705.05363}, 2017.

\bibitem[Senechal et~al.(2015)Senechal, McDuff, and
  Kaliouby]{senechal2015facial}
Thibaud Senechal, Daniel McDuff, and Rana Kaliouby.
\newblock Facial action unit detection using active learning and an efficient
  non-linear kernel approximation.
\newblock In \emph{Proceedings of the IEEE International Conference on Computer
  Vision Workshops}, pages 10--18, 2015.

\bibitem[van Schaik and Burkart(2011)]{van2011social}
Carel~P van Schaik and Judith~M Burkart.
\newblock Social learning and evolution: the cultural intelligence hypothesis.
\newblock \emph{Philosophical Transactions of the Royal Society B: Biological
  Sciences}, 366\penalty0 (1567):\penalty0 1008--1016, 2011.

\end{thebibliography}

\end{document}